\documentclass[sigconf,nonacm]{acmart}

\usepackage{multirow}
\usepackage{booktabs}
\usepackage[table]{xcolor}
\usepackage{graphicx}
\usepackage{subcaption}
\usepackage[normalem]{ulem}
\usepackage[ruled, linesnumbered]{algorithm2e}

\usepackage{pifont}

\definecolor{maroon}{cmyk}{0,0.87,0.68,0.32}
\definecolor{bamboo}{cmyk}{0.4,0,0.3,0}
\definecolor{apple}{cmyk}{0.41,0.4,0.76,0}
\definecolor{jialingshui}{cmyk}{0.47,0,0.49,0}
\definecolor{sea}{cmyk}{1,0.67,0.16,0.03}

\AtBeginDocument{%
  }

\setcopyright{acmlicensed}
\copyrightyear{2026}
\acmYear{2026}
\acmDOI{XXXXXXX.XXXXXXX}
\acmISBN{978-1-4503-XXXX-X/2018/06}

\acmSubmissionID{2310}

\begin{document}

\title{Semi-Supervised Flow Matching for Mosaiced and Panchromatic Fusion Imaging}


\author{Peiming Luo}
\authornote{Equal Contribution.}
\affiliation{%
  \institution{Southeast University}
  \city{Nanjing}
  \country{China}}
\email{peiming_luo@seu.edu.cn}

\author{Nan Wang}
\authornotemark[1]
\affiliation{%
  \institution{Hunan University}
  \city{Changsha}
  \country{China}}
\email{wangn@hnu.edu.cn}

\author{Litong Liu}
\affiliation{%
  \institution{Southeast University}
  \city{Nanjing}
  \country{China}}
\email{lliu0118@student.monash.edu}

\author{Jiahan Huang}
\affiliation{%
  \institution{Southeast University}
  \city{Nanjing}
  \country{China}}
\email{jiahan_huang@seu.edu.cn}

\author{Chenxu Wu}
\affiliation{%
  \institution{Southeast University}
  \city{Nanjing}
  \country{China}}
\email{220250763@seu.edu.cn}

\author{Renwei Dian}
\authornotemark[2]
\affiliation{%
  \institution{Hunan University}
  \city{Changsha}
  \country{China}}
\email{drw@hnu.edu.cn}

\author{Junming Hou}
\authornote{Corresponding authors.}
\affiliation{%
  \institution{Southeast University}
  \city{Nanjing}
  \country{China}}
\email{junming_hou@seu.edu.cn}

\renewcommand\footnotetextcopyrightpermission[1]{}

\renewcommand{\shortauthors}{Peiming Luo et al.}

\begin{abstract}
Fusing a low resolution (LR) mosaiced hyperspectral image (HSI) with a high resolution (HR) panchromatic (PAN) image 
offers a promising avenue for video-rate HR-HSI imaging via single-shot acquisition, yet its severely ill-posed nature remains a significant challenge.
In this work, we propose a novel semi-supervised flow matching framework for mosaiced and PAN image fusion. Unlike previous diffusion-based approaches constrained by specific protocols or handcrafted assumptions, our method seamlessly integrates an unsupervised scheme with flow matching, resulting in a generalizable and efficient generative framework. 
Specifically, our method follows a two-stage training pipeline. First, we pretrain an unsupervised prior network to produce an initial pseudo HR-HSI.
Building on this, we then train a conditional flow matching model to generate the target HR-HSI, introducing a random voting mechanism that iteratively refines the initial HR-HSI estimate, enabling robust and effective fusion.
During inference, we employ a conflict-free gradient guidance strategy that ensures spectrally and spatially consistent HR-HSI reconstruction.
Experiments on multiple benchmark datasets demonstrate that our method achieves superior quantitative and qualitative performance by a significant margin compared to representative baselines. Beyond mosaiced and PAN fusion, our approach provides a flexible generative framework that can be readily extended to other image fusion tasks and integrated with unsupervised or blind image restoration algorithms. 

\end{abstract}



\begin{teaserfigure}
	\centering
	\includegraphics[width=0.98\textwidth]{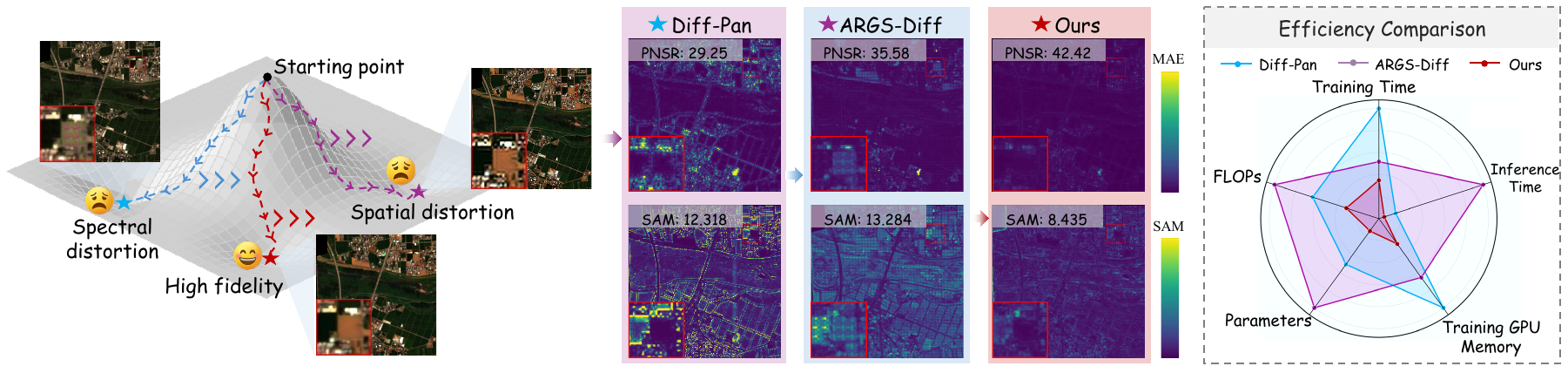}
	\caption{Left: Comparison between cutting-edge diffusion models and our approach. Unlike existing approaches that often get trapped in sub-optimal solutions, our method delivers superior spatial and spectral fidelity. Middle: Our method significantly outperforms representative supervised (Diff-Pan \protect\cite{cao2024diffusion}) and unsupervised (ARGS-Diff \protect\cite{zhu2025self}) diffusion-based approaches in both quantitative and qualitative evaluations. Right: Our method achieves clear advantages in computational efficiency.}
	\label{graphic abstract}
\end{teaserfigure}
\maketitle
\thispagestyle{plain}
\pagestyle{plain}

\section{Introduction}
Hyperspectral imaging captures rich spectral information beyond human visual perception, underpinning critical applications in remote sensing \cite{li2019deep}, medical diagnostics \cite{backman2000detection, hadoux2019non}, industrial inspection \cite{cui2025chip, hong2026hyperspectral}, etc. However, traditional scanning-based hyperspectral imaging approaches suffer from poor temporal resolution, rendering them largely unsuitable for dynamic scenes. To circumvent this limitation, mosaic-based snapshot hyperspectral imaging, inspired by the RGB Bayer filter array, has emerged as a highly practical alternative. By employing on-sensor spectral filter array (SFA), this paradigm captures 2D spatial-spectral projections in a single shot, thereby enabling video-rate hyperspectral image (HSI) reconstruction \cite{liu2024super, nature_bian, mu2026hyperspectral}. Although temporally efficient, the mosaic-based hyperspectral imaging technique inherently trades spatial resolution for the ability to capture multiple spectral bands.

\begin{figure}[!t]
    \centering
    \includegraphics[width=0.98\columnwidth]{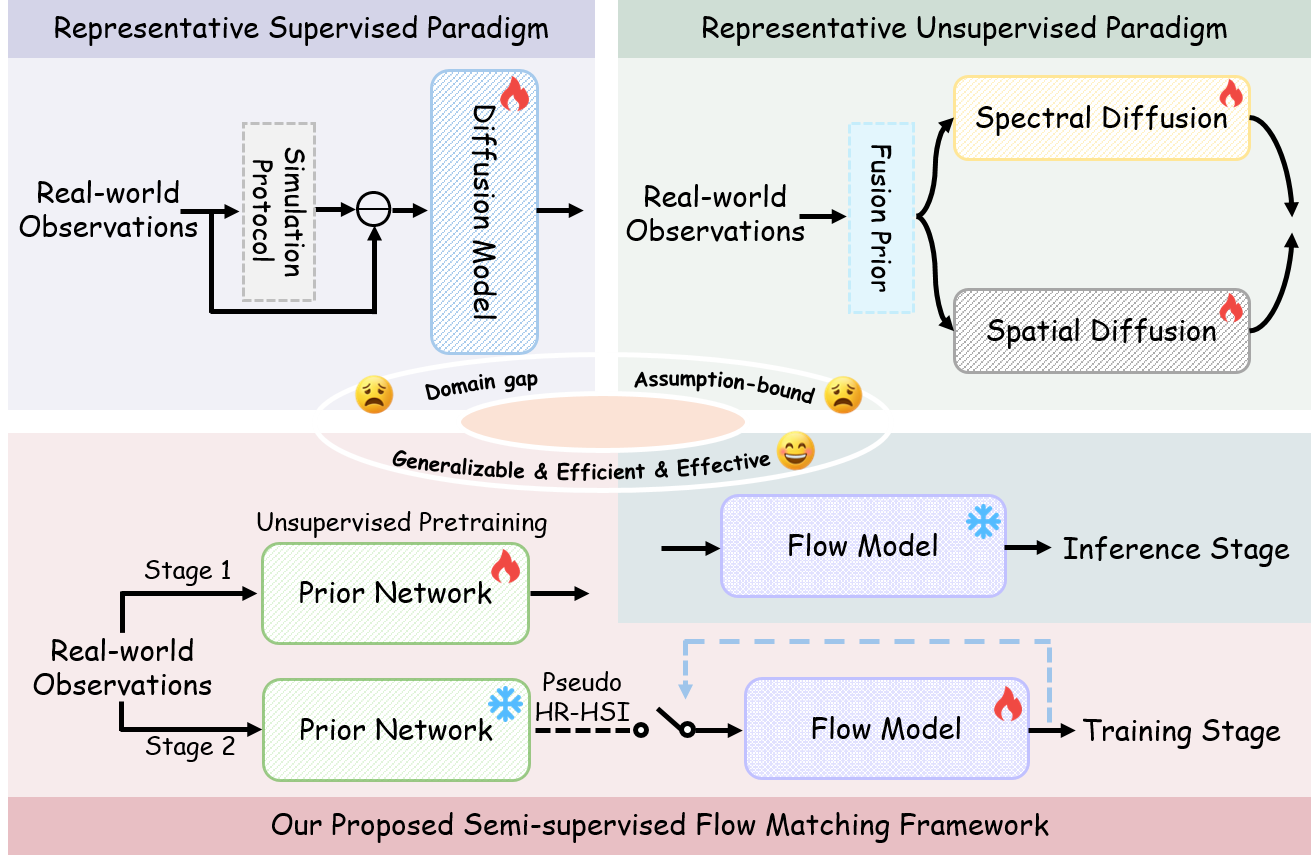}
    \caption{Schematic comparison of existing supervised/unsupervised diffusion models and our proposed semi-supervised flow matching framework. Supervised diffusion models are often trained on protocol-specific simulation data (\emph{e.g.}, Wald’s protocol), introducing a noticeable domain gap in real-world scenarios; while unsupervised approaches rely on predefined assumptions, such as low-rank tensor decomposition and subspace projection, leading to limited flexibility and generalizability. Instead, our method unifies unsupervised pretraining with generative modeling, offering a flexible, task-agnostic framework.
    }
    \label{fig:toy}
\end{figure}

A compelling strategy to mitigate this limitation is to fuse the low-resolution (LR) mosaiced image with a high-resolution (HR) panchromatic (PAN) image of the same scene, thereby leveraging their complementary spectral and spatial characteristics. An intuitive solution for the mosaiced and PAN image fusion is to decompose the task into two sub-tasks, i.e., spectral demosaicing and pansharpening. That is, the mosaiced image is firstly used to predict a LR HSI with a fully-defined number of channels, which still exhibits a $2\times2$ spatial resolution gap with the PAN image. Then, the LR HSI is spatially sharpened by the HR PAN to obtain the HR-HSI, thereby yielding a favorable fusion performance. Existing approaches for demosaicing and pansharpening can be broadly categorized into traditional methods \cite{mihoubi2017multispectral, tsagkatakis2018graph, zhang2021polarization, wen2021sparse} and deep learning (DL)-based methods \cite{habtegebrial2019deep, SFNet, VBPN, PanGAN, liu2025exploring, huang2025wavelet, EFN}. Traditional techniques typically rely on hand-crafted priors \cite{mihoubi2017multispectral}, such as matrix factorization \cite{tsagkatakis2018graph}, tensor decomposition \cite{zhang2021polarization}, and sparse coding \cite{wen2021sparse}. While mathematically rigorous, these methods often struggle to model complex spectral-spatial correlations due to limited representation capacity, or they suffer from high computational costs, which constrains their overall performance. In contrast, DL-based methods have demonstrated superior performance by learning end-to-end mappings from paired training data. These can be further divided into supervised and unsupervised paradigms. Supervised approaches \cite{he2024lgct, SFNet, fang2025content, he2025adaptive, VBPN, xu2025laboring} necessitate ground-truth HR-HSI for training. However, acquiring such ground truth in real-world scenarios is notoriously difficult and cost-prohibitive. To circumvent this bottleneck, unsupervised frameworks \cite{PanGAN, chi2025cross, du2026unsupervised, xu2025nonlinear, EFN} have emerged as a promising alternative by eliminating the dependency on ground-truth HSI. Notably, Wang \textit{et al.} recently proposed an unsupervised equivariant fusion network (EFN) \cite{EFN}, which aims to address this task through a unified, one-step architecture.

Recently, diffusion models have gained increasing attention in hyperspectral and multispectral image fusion tasks, significantly outperforming conventional DL-based approaches. In general, these methods mainly operate within either supervised or unsupervised paradigms. In supervised settings, diffusion models are trained to learn the residual between high-resolution reference images and their simulated low-resolution counterparts, typically following protocols such as Wald’ s protocol~\cite{cao2024diffusion,hou2026nodiff}. While these methods achieve impressive fusion results on synthetic data, their reliance on predefined simulation strategies introduces a notable domain gap when applied to actual observations, thereby limiting their applicability in real-world scenarios.
In contrast, unsupervised diffusion models often leverage low-rank tensor decomposition subspace projection techniques to disentangle the target hyperspectral image into spatial and spectral components, which then serve as the modeling objectives for the diffusion process~\cite{pang2024hir,zhu2025self,xiao2025hyperspectral,xu2026coupled}. Although this paradigm is mathematically elegant and offers strong generalization potential in principle, its practical applicability is inherently constrained by the assumption that such decompositions can be reliably derived from the observed data.
In real-world scenarios, however, the spatial and spectral components are not explicitly available and may be corrupted by sensor noise, spectral variability, or complex mixing effects, making accurate decomposition challenging. Consequently, methods that rely on fixed or precomputed decompositions face significant limitations when applied to real-world data.

In addition, despite their proven efficacy, diffusion models have yet to be explored for mosaiced and PAN image fusion. On the one hand, the unavailability of high-resolution reference images in real-world scenarios precludes supervised training. On the other hand, designing an appropriate and reliable decomposition principle tailored to mosaiced data remains challenging, limiting the applicability of existing unsupervised frameworks. To address these limitations, we propose a general and effective generative framework for mosaiced and PAN image fusion, which achieves superior spatial and spectral fidelity compared to state-of-the-art representative diffusion-based baselines, as illustrated in Fig.~\ref{graphic abstract}. To be specific, the proposed framework is implemented via a two-stage training pipeline. First, we pretrain a lightweight prior fusion network in an unsupervised manner to produce pseudo HR-HSI. Subsequently, we compute the residual between the
obtained pseudo HR-HSI and channel-wise expanded PAN image, serving as the training target for the flow matching model. During this phase, the pseudo HR-HSI is iteratively updated using the predictions of the flow matching model via a simple yet effective random voting mechanism. Fig.~\ref{fig:toy} compares the training paradigms of our proposed semi-supervised flow matching framework with those of previous supervised and unsupervised diffusion models.
The main contributions of this work are summarized as follows:
\begin{itemize}
    \item We propose a generalizable and efficient semi-supervised flow matching framework that achieves superior fusion performance and computational efficiency, significantly outperforming state-of-the-art approaches.
    \item We introduce a novel and effective random voting mechanism, enabling robust training of the flow matching model directly on real-world data, rather than being constrained by specific simulation protocols or predefined assumptions as in existing diffusion-based approaches.
    \item We employ a conflict-free gradient guidance mechanism during inference to mitigate conflicts among multiple objective terms, providing a unified gradient signal to steer the generation toward desired spatial-spectral fidelity.
    \item Beyond mosaiced and PAN image fusion, this work establishes a flexible and extensible paradigm for real-world image restoration, which can be readily integrated with existing unsupervised and blind methods.
\end{itemize}

\section{Related Work}
\subsection{Traditional Fusion Algorithms}
As mentioned above, the mosaiced and PAN image fusion can be decomposed into the spectral demosaicing and pansharpening. Early spectral demosaicing methods \cite{mihoubi2017multispectral, tsagkatakis2018graph, zhang2021polarization, wen2021sparse} predominantly relied on hand-crafted priors. A representative classical approach is the PPID algorithm \cite{mihoubi2017multispectral}, which achieves robust performance by leveraging a pseudo-panchromatic image estimated directly from the mosaiced input. In parallel, traditional pansharpening techniques exploited priors such as spectral-spatial consistency, sparse representation, and structural similarity \cite{GSA, BDSD, MTF_GLP, SFIM, CNMF, HySure}. However, these methods are often constrained by rigid hand-crafted priors, resulting in noticeable spectral or spatial distortions. Furthermore, most classical approaches rely on computationally intensive iterative optimization schemes, rendering them inefficient and impractical for real-world deployment.

\subsection{Deep Learning-based Fusion Approaches}
Conversely, DL-based approaches deliver impressive fusion results owing to their powerful non-linear approximating capability, while maintaining high computational efficiency through streamlined single-pass inference. For instance, Liu \textit{et al.} \cite{liu2025exploring} proposed polarity memory network (PMNet) with quant attention to establish global correlation, thus reconstructing high-quality hyperspectral image from mosaiced image. Zhang \textit{et al.} \cite{VBPN} designed a deep variational pansharpening network, namely VBPN, to yield superior results with enhanced interpretability and generalization. Huang \textit{et al.} proposed wavelet-assisted fusion network (WFANet) for pansharpening \cite{huang2025wavelet}, aiming to enable lossless reconstruction. However, these supervised methods rely heavily on ground-truth HSI, which is usually inaccessible in the real-world scenario. Recent research has increasingly pivoted towards unsupervised paradigms. For example, Ma \textit{et al.} \cite{PanGAN} proposed a GAN-based framework (PanGAN) that enforces explicit cycle-consistency constraints to preserve both spectral and spatial fidelity. In parallel, Feng \textit{et al.} \cite{LSAN} integrated an equivariant imaging prior into an unsupervised demosaicing architecture. More recently, Wang \textit{et al.} proposed the equivariant fusion network \cite{EFN} for the mosaiced and PAN image fusion, demonstrating promising results in unsupervised settings.

\subsection{Diffusion Models for Hyperspectral and Multispectral Image Fusion}
In recent years, diffusion models~\cite{ho2020denoising,song2021denoising} have attracted considerable attention across a variety of vision tasks. 
In hyperspectral and multispectral image fusion, a growing number of diffusion-based models have been proposed, significantly advancing fusion performance compared to conventional DL-based approaches. In general, these methods are implemented via two primary training paradigms: supervised approaches and unsupervised frameworks.
Technically, in supervised settings, models are trained to learn the residual between high-resolution reference images and their low-resolution counterparts~\cite{cao2024diffusion,kim2025u,hou2026nodiff}, where these image pairs are typically simulated following certain protocols such as Wald’s protocol. Therefore, these methods inevitably lead to a significant domain gap when applied to real-world scenarios.
In unsupervised diffusion frameworks, models often rely on certain priors, such as low-rank tensor decomposition or subspace projection techniques~\cite{pang2024hir,zhu2025self,xiao2025hyperspectral,xu2026coupled}, to disentangle the target hyperspectral image into spatial and spectral components, which are then used as objectives for diffusion modeling. While mathematically elegant and theoretically capable of strong generalization, its practical effectiveness is
fundamentally limited by the assumption that such decompositions can be reliably obtained from the observed data. In real-world scenarios, however, the spatial and spectral components are rarely directly accessible and are often degraded by factors such as sensor noise, spectral variability, or complex mixing effects, making accurate decomposition highly challenging. As a result, methods dependent on fixed or precomputed decompositions exhibit limited flexibility and generalizability in practical applications, such as mosaic-based hyperspectral imaging. To address these challenges, we propose a flexible and effective semi-supervised flow matching framework that can be readily integrated with existing unsupervised approaches or blind restoration techniques. 

\begin{figure*}[!t]
    \centering
    \includegraphics[width=0.95\textwidth]{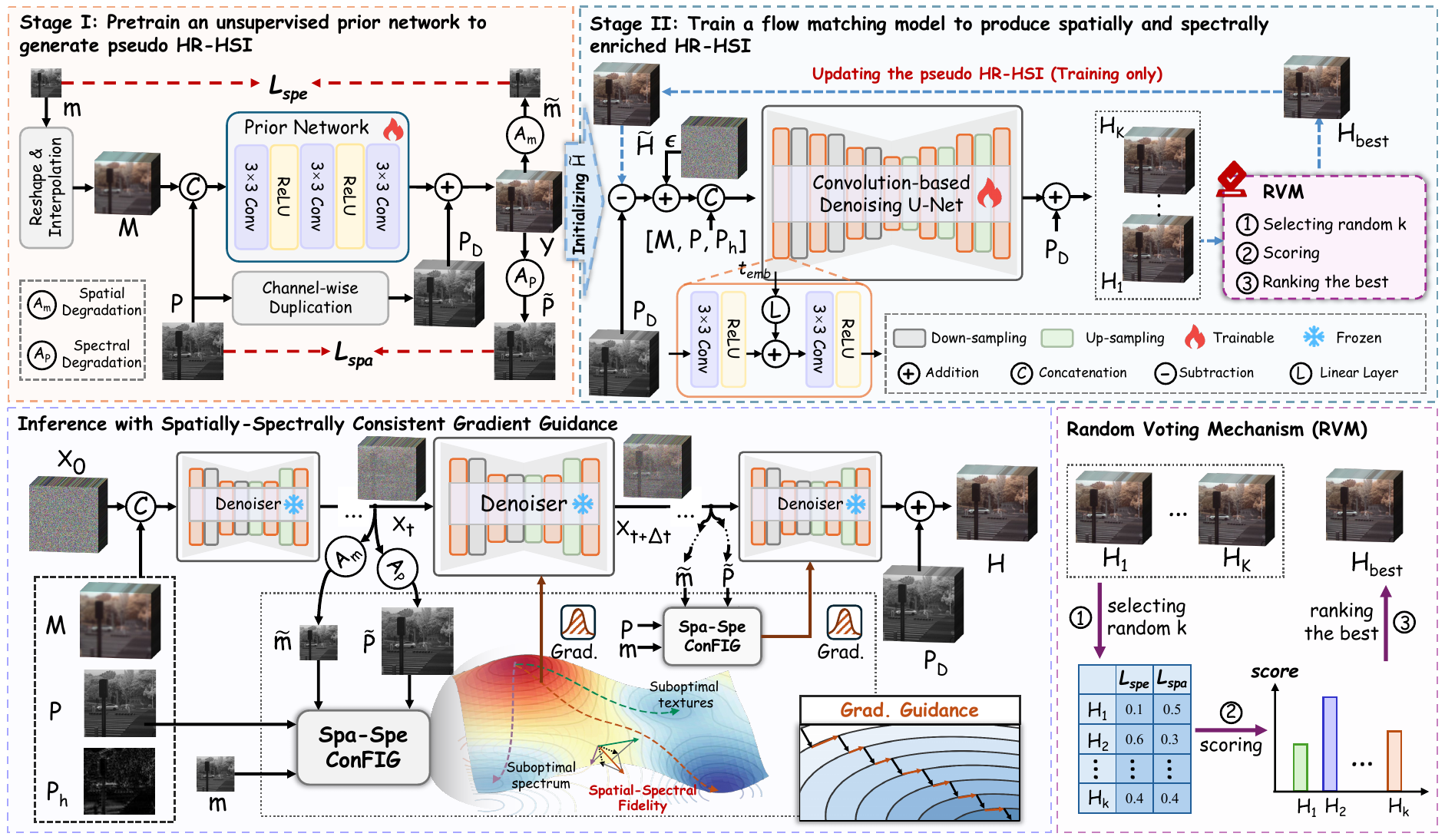}
    \caption{Framework overview of the proposed semi-supervised flow matching framework for mosaiced and PAN image fusion, implemented via a two-stage training pipeline. First, a prior network is pretrained in an unsupervised manner to generate pseudo HR-HSI estimate.  Subsequently, a flow matching model is trained to generate the desired HR-HSI, where the residual between the pseudo HR-HSI and channel-expanded PAN image serves as the initial training target and is iteratively updated through a random voting mechanism. During inference, the trained model directly generate an HR-HSI with superior spatial and spectral fidelity, guided by a conflict-free gradient guidance strategy.}
    \label{framework}
\end{figure*}
 
\section{Methodology}
In this section, we present our framework for mosaiced and PAN image fusion, organized into four parts: 1) Problem Formulation and Pretraining (Sec.~\ref{method:PFP}); 2) Conditional Flow Matching Model (Sec.~\ref{method:CFM}); 3) Random Voting Mechanism (Sec.~\ref{method:RVM}); and 4) Conflict-free Guidance Sampling (Sec.~\ref{method:CGS}).
\subsection{Problem Formulation and Pretraining}\label{method:PFP}
\noindent\textbf{Problem Formulation.} Let $\mathbf{m} \in \mathbb{R}^{1 \times S}$ and $\mathbf{P} \in \mathbb{R}^{1 \times N}$ denote the observed LR mosaiced and HR PAN images, where $S$ and $N$ represent the spatial dimensions. Our goal is to reconstruct the target HR-HSI $\mathbf{H} \in \mathbb{R}^{{c} \times {N}}$, where $c$ is the number of spectral bands. The degradation process of the mosaiced image is defined by a fixed spatial degradation function $\mathcal A_m(\cdot)$, which comprises an $8 \times 8$ mosaicing and $2 \times 2$ average downsampling. 
In contrast, the PAN image is modeled via a learnable spectral degradation operator $\mathcal A_P(\cdot)$, as the SRF is typically unavailable in practice.

\noindent\textbf{Pretraining.}
We propose a two-stage training pipeline that enables robust training of the flow matching model directly on real-world data, overcoming the limitations of existing supervised and unsupervised diffusion-based approaches. As illustrated in Fig.~\ref{framework}, we first pretrain a prior network $\mathcal G_{\theta}$ in an unsupervised manner to fuse the interpolated mosaiced image $\mathbf{M} \in \mathbb{R}^{c \times N}$ and the PAN image, producing the initial HR-HSI estimate $\mathbf{Y} = \mathcal{G}_{\theta}(\mathbf{P}, \mathbf{M})$. To guide $\mathcal{G}_{\theta}$ towards the correct solution space without paired data, we employ an Equivariant Imaging (EI) framework following~\cite{EFN}. Specifically, this pretraining stage is optimized through a spatial-spectral observation consistency loss and an equivariant loss:
\begin{equation} \label{eq1}
\mathcal{L}_{pre} = \mathcal{L}_{oc}(\theta) + \mathcal{L}_{ei}(\theta),
\end{equation}
where $\mathcal{L}_{oc}$ ensures fidelity to the original inputs, defined as:
\begin{equation} \label{eq2}
\mathcal{L}_{oc}(\theta) = \left\| \mathbf{P} - \mathcal{A}_P(\mathbf{Y}) \right\|_2^2 + \left\| \mathbf{m} - \mathcal{A}_m(\mathbf{Y}) \right\|_2^2.
\end{equation}
Furthermore, $\mathcal{L}_{ei}$ enforces transformation equivariance. Given a spatial transformation operator $\mathcal{T}(\cdot)$ (\emph{e.g.}, random flipping or rotation), the equivariant loss is defined as:
\begin{equation} \label{eq3}
\mathcal{L}_{ei}(\theta) = \left \| \mathcal{T}(\mathbf{Y}) - \mathcal{G}_{\theta}(\mathcal{A}_P(\mathcal{T}(\mathbf{Y})), \mathcal{A}_m(\mathcal{T}(\mathbf{Y}))) \right \|_2^2.
\end{equation}
The optimized $\mathbf{Y}$ serves as the pseudo HR-HSI used for the subsequent training of the flow matching model.

\subsection{Conditional Flow Matching Model}\label{method:CFM}
Building on the pretrained prior network $\mathcal{G}_{\theta}$, we train the flow matching model in a manner akin to supervised learning, targeting the residual between the HR-HSI and the channel-expanded PAN image $\mathbf{P}_{D} \in \mathbb{R}^{c \times N}$. The initial target residual is defined as $\mathbf {\widetilde{R}} = \mathbf{Y} - \mathbf{P}_{D}$. 
Notably, the target residual is not fixed but evolves during training, in contrast to conventional supervised diffusion models where the residual remains static.
This is because the initial estimate $\mathbf{Y}$ is dynamically updated using the predictions $\mathbf{\widetilde{H}}$ (initialized as $\mathbf{\widetilde{H}} = \mathbf{Y}$) of the flow matching model via a random voting mechanism. 
Consequently, the flow matching model is trained to predict an evolving target residual, denoted as $ \mathbf{\widetilde{R}} = \mathbf{\widetilde{H}} - \mathbf{P}_{D}$.

To model this evolving distribution, we employ conditional flow matching (CFM), which learns a continuous-time vector field that transports a simple prior distribution to the target data distribution~\cite{lipman2023flow}. Let $\mathbf{X}_0 \sim \mathcal{N}(0, I)$ denote the initial noise and $\mathbf{X}_1 = \mathbf{\widetilde{R}}$ represent the target residual that is progressively updated. The CFM framework defines a probability density path along with a corresponding target vector field $\mathcal{V}_t(\mathbf{X}_t)$. A purely convolutional U-Net architecture is then trained to predict this vector field over continuous time $t \in [0, 1]$, conditioned on the spatial and spectral features $\mathbf{C}$ extracted from $\mathbf{M}$, $\mathbf{P}$, and $\mathbf{P}_{h}$, where $\mathbf{P}_{h}$ represents the high-frequency PAN details filtered by a Laplacian operator. To further constrain the flow matching process while enforcing spatial and spectral consistency, we leverage a joint loss function that integrates velocity matching on both direct and transformed inputs with degradation penalties. Specifically, the direct flow matching loss is defined as: 
\begin{equation} \label{eq4}
\mathcal{L}_{dir} = \mathbb{E}_{t, \mathbf{X}_0, \mathbf{X}_1} [\left \| \mathcal{V}_{\phi}(\mathbf{X}_t, t, \mathbf{C}) - \mathcal{V}_t(\mathbf{X}_t) \right\|_2^2],
\end{equation}
where $\mathcal{V}_{\phi}(\mathbf{X}_t, t, \mathbf{C})$ denotes the predicted conditional vector field.

By applying the random spatial transformation $\mathcal{T}(\cdot)$ to the noised residual, we enforce spatial consistency in the predicted vector field. The corresponding flow matching loss is defined as:

\begin{equation} \label{eq5}
\mathcal{L}_{trans} = \mathbb{E}_{t, \mathcal{T}(\mathbf{X}_0), \mathcal{T}(\mathbf{X}_1)} [\left \|\mathcal{V}_{\phi}(\mathcal{T}(\mathbf{X}_t), t, \mathcal{T}(\mathbf{C})) - \mathcal{V}_t(\mathcal{T}(\mathbf{X}_t)) \right \|_2^2].
\end{equation}
Then, we add the predicted residual back to $\mathbf{P}_{D}$, generating the desired HR-HSI.Additionally, we compute the degradation consistency loss using the learnable spectral degradation $\mathcal{A}_{P}(\cdot)$ (initialized from the pretraining stage) and the fixed spatial degradation $\mathcal{A}_{m}(\cdot)$:
\begin{equation} \label{eq6}
\mathcal{L}_{deg} = \lambda (\left \| \mathbf{P} - \mathcal{A}_P({\mathbf{H}})\right \|_2^2 + \left\|\mathbf{m} - \mathcal{A}_m({\mathbf{H}}) \right \|_2^2),
\end{equation}
where $\lambda$ is a weight parameter applied after a defined warmup period. The total training loss for the CFM model is formulated as:
\begin{equation} \label{eq7a}
\mathcal{L}_{total} = \mathcal{L}_{dir} + \mathcal{L}_{trans} + \mathcal{L}_{deg}.
\end{equation}

\subsection{Random Voting Mechanism}\label{method:RVM}
Unlike conventional supervised diffusion models, which are typically trained to learn the static residual between a high-resolution reference image and its simulated low-resolution counterpart, our approach operates directly on real-world data where ground truth is unavailable. To this end, we introduce a dynamic voting mechanism that progressively updates $\mathbf{\widetilde{H}}$ during training, where $\mathbf{\widetilde{H}}$ is initialized by the output of the pretrained prior network.

Specifically, within a fixed evaluation window of $K$ epochs, we randomly select $k$ checkpoints of the CFM model to predict a set of candidate HR-HSI $\{\mathbf{H}_i\}_{i=1}^k$. For each candidate prediction, we evaluate its spectral and spatial observation consistency criterion. Following the formulation of the degradation consistency loss, we define an evaluation function $\mathcal{E}(\cdot)$ for any given estimate $\mathbf{X}$ by degrading it back to the observation space:
\begin{equation} \label{eq7b}
\mathcal{E}(\mathbf{X}) = \left\|\mathbf{P} - \mathcal{A}_P(\mathbf{X})\right\|_2^2 + \left\|\mathbf{m} - \mathcal{A}_m(\mathbf{X})\right\|_2^2 .
\end{equation}

\begin{figure}[!t]
    \centering
    \includegraphics[width=0.98\columnwidth]{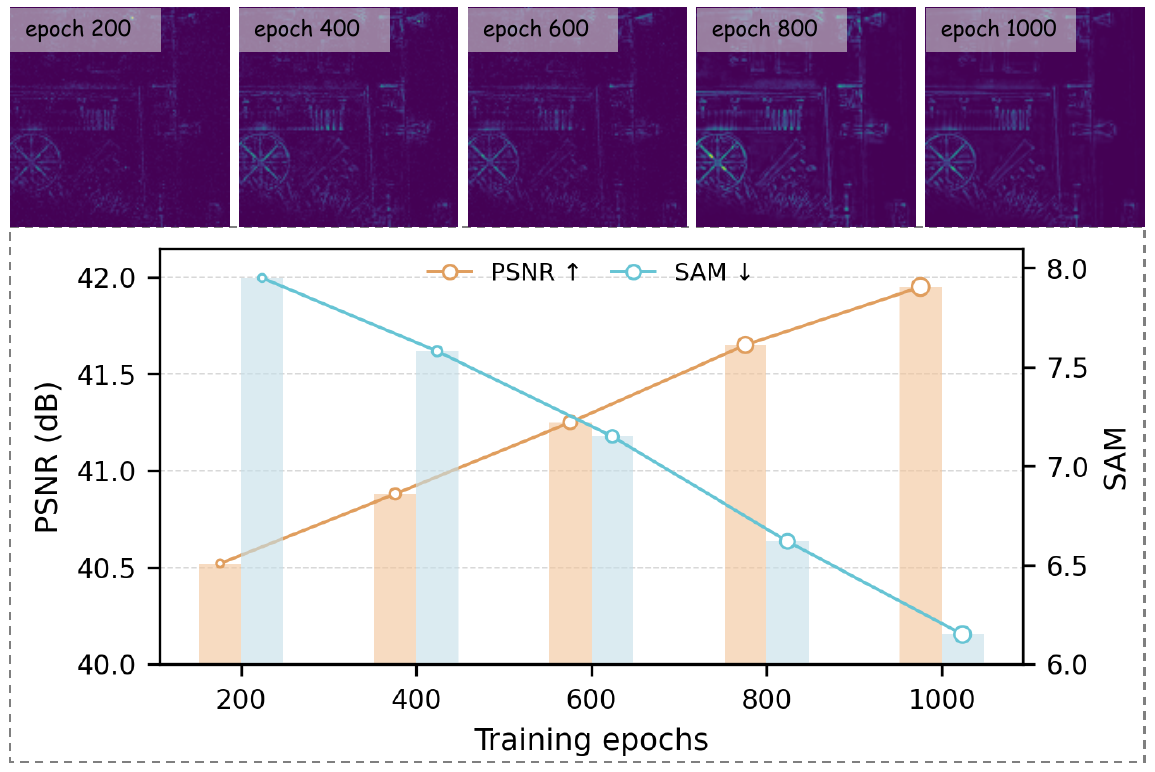}
    \caption{Variation of the target residual, alongside the spatial and spectral metrics of the updated HR-HSI (illustrated on a simulated example). These progressive refined results clearly validate the effectiveness of the random voting mechanism. }
    \label{fig:RVM}
\end{figure}

We then assess the win rate of the model predictions against the current $\mathbf{\widetilde{H}}$. If the candidate models outperform the current $\mathbf{\widetilde{H}}$ in at least a proportion $p$ of the sampled epochs:
\begin{equation} \label{eq7c}
\frac{1}{k} \sum_{i=1}^k \mathbb{I} \left( \mathcal{E}(\mathbf{H}_i) < \mathcal{E}(\mathbf{\widetilde{H}}) \right) \ge p ,
\end{equation}
where $\mathbb{I}(\cdot)$ denotes the indicator function, and $p$ is a predefined threshold (set to 0.75 in our experiments). The updated $\mathbf{\widetilde{H}}$ is selected with the minimum spatial and spectral deviations:
\begin{equation}
\mathbf{\widetilde{H}} \leftarrow \mathbf{H}_{best} = \arg\min_{\mathbf{{H}}_i} \mathcal{E}(\mathbf{{H}}_i).
\end{equation}
This closed-loop voting strategy enables progressive refinement of the training targets, allowing the flow matching model to learn from increasingly reliable supervision, while mitigating the risk of collapsing to a suboptimal solution caused by inaccuracies in the early-stage pseudo-label. Fig.~\ref{fig:RVM} illustrates the training dynamics, with progressively refined residual along with improved spatial and spectral metrics, consistently validating its effectiveness.


\subsection{Conflict-free Guided Sampling}\label{method:CGS}
Recently, gradient guidance mechanism~\cite{chung2023diffusion} has been widely adopted in low-level vision tasks~\cite{lin2025aglldiff,li2025difiisr}, where gradient signals derived from reference-free attribute constraints are used to iteratively steer the reverse diffusion process toward high-quality restoration. This property makes it particularly suitable for real-world scenarios without ground-truth supervision. Motivated by this, we integrate gradient guidance at inference time to enforce spatial and spectral fidelity. At each step $t$ of the flow matching process, the current state $\mathbf{X}_t$ serves as an intermediate progressive estimate of the target residual. We construct the intermediate HR-HSI prediction by adding it to $\mathbf{P}_D$. Next, we compute the spatial degradation loss $\mathcal{L}_{spa}$ and the spectral degradation loss $\mathcal{L}_{spe}$:
\begin{equation}
\begin{split}
&\mathcal{L}_{spa} = \left\| \mathbf{m} - {\mathcal{A}_m}(\mathbf{X}_t+\mathbf{P}_D)\right\|_2^2, \\
&\mathcal{L}_{spe} = \left\| \mathbf{P} - {\mathcal{A}_P}(\mathbf{X}_t+\mathbf{P}_D)\right\|_2^2.
\end{split}
\end{equation}
Let $\mathbf{g}_{spa} = \nabla_{\mathbf{X}_t} \mathcal{L}_{spa}$ and $\mathbf{g}_{spe} = \nabla_{\mathbf{X}_t} \mathcal{L}_{spe}$ denote the gradients of the two objectives with respect to $\mathbf{X}_t$. Previous guidance strategies typically combine these gradients via direct summation. However, this introduces the risk of gradient conflicts (\emph{i.e.}, $\mathbf{g}_{spa}^\top \mathbf{g}_{spe} < 0$) during optimization, resulting in oscillatory trajectories or biasing the generation toward a certain attribute (\emph{e.g.}, spatial or spectral).

\begin{figure}[!t]
    \centering
    \includegraphics[width=0.98\columnwidth]{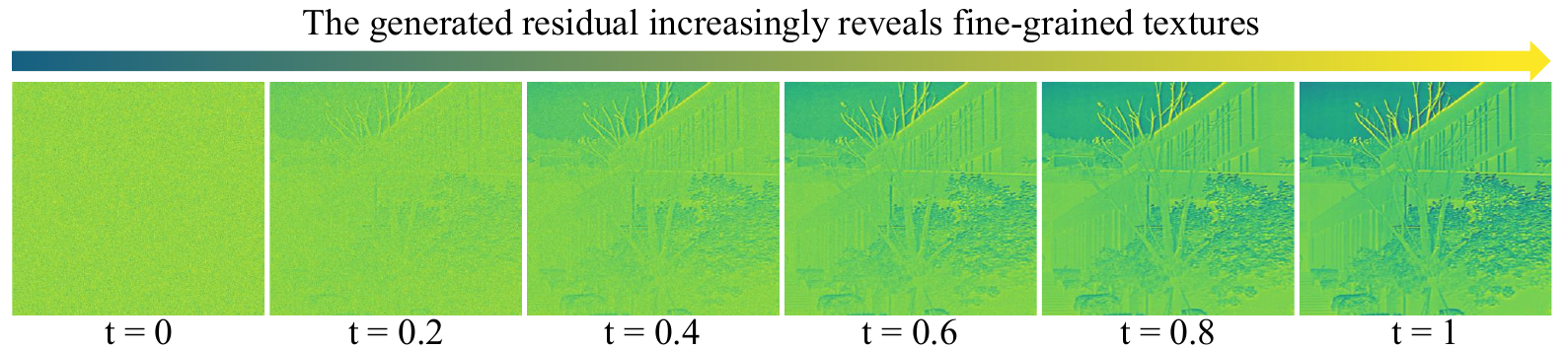}
    \caption{Variation of the target residual during the sampling process (visualized on a real-world example).
    }
    \label{fig:sampling residual}
\end{figure}

Drawing the inspiration from the principles of partial differential equations (PDEs)~\cite{liu2025config}, we adopt a conflict-free strategy to unify the spatial and spectral gradient updates:
\begin{equation}
\begin{split}
&\mathbf{g}_{update} = (\mathbf{g}_{spa}^\top \mathbf{g}_v + \mathbf{g}_{spe}^\top \mathbf{g}_v)\mathbf{g}_v, \\
&\mathbf{g}_v = \mathcal{U} \left[ \mathcal{U}(\mathcal{O}(\mathbf{g}_{spa}, \mathbf{g}_{spe})) + \mathcal{U}(\mathcal{O}(\mathbf{g}_{spe}, \mathbf{g}_{spa})) \right],
\end{split}
\end{equation}
where $\mathcal{O}(\mathbf{g}_{spa}, \mathbf{g}_{spe}) = \mathbf{g}_{spe} - \frac{\mathbf{g}_{spa}^\top \mathbf{g}_{spe}}{\left\|\mathbf{g}_{spa}\right\|_2^2} \mathbf{g}_{spa}$ denote the orthogonality operator, and $\mathcal{U}(\mathbf{g}) = \mathbf{g} / \left\|\mathbf{g}\right\|_{1}$ normalizes a vector to unit length. This approach adaptively aligns the spatial and spectral gradients, eliminating the risk of gradient conflicts and preventing one objective from dominating the optimization. The resulting update direction $\mathbf{g}_{update}$ enables spatially and spectrally consistent optimization, \emph{i.e.}, $\mathbf{g}_{update}^\top \mathbf{g}_{spa} > 0$ and $\mathbf{g}_{update}^\top \mathbf{g}_{spe} > 0$. 
Finally, we incorporate this conflict-free gradient guidance into the flow matching sampling process. Following the guided flow matching formulation, the modified update step for the ODE solver is defined as:
\begin{equation}
\mathbf{X}_{t+\Delta t} = \mathbf{X}_t + \Delta t \cdot \mathcal{V}_{\phi}(\mathbf{X}_t, t, \mathbf{C}) - \gamma \cdot \mathbf{g}_{update},
\end{equation}  
where $\gamma$ controls the gradient guidance. The velocity field $\mathcal{V}_{\phi}$ drives the distribution-matching trajectory along the prior, 
while the guidance term $\gamma \cdot \mathbf{g}_{update}$ enforces a conflict-free descent direction on the data manifold, guiding the generation toward a balanced spatial and spectral fidelity. As depicted in Fig.~\ref{fig:sampling residual}, the target residual undergoes progressive refinement throughout sampling, exhibiting finer textures and demonstrating the model's efficacy.

\begin{table*}[t]
\centering
\label{tab:ablation}
\renewcommand\arraystretch{1.0}
\resizebox{1.0\textwidth}{!}{%
\begin{tabular}{l|cccc|cccc|ccc}
\hline
\rowcolor[HTML]{ECF4FF} 
&\multicolumn{4}{c|}{\textbf{CAVE}}&\multicolumn{4}{c|}{\textbf{Chikusei}}&\multicolumn{3}{c}{\textbf{Real-world}}\\
    \rowcolor[HTML]{ECF4FF} 
    \textbf{Method}& PSNR$\uparrow$  & SSIM$\uparrow$  & SAM$\downarrow$ & ERGAS$\downarrow$   & PSNR$\uparrow$  & SSIM$\uparrow$  & SAM$\downarrow$ & ERGAS$\downarrow$  & QNR$\uparrow$   & D$_\lambda$$\downarrow$ & D$_S$$\downarrow$      \\ \hline
            
    PPID-PanGAN  & 34.85 & 0.9457 & 10.8829 & 4.0853 & 33.25  & 0.8988 & 23.2011 & 8.3582 & 0.8476  & 0.0323 & 0.0432  \\
    PPID-VBPN    & 33.21 & 0.9007 & 10.7391 & 4.7555 & 33.37  & 0.8800 & 21.8142 & 8.1733 & 0.7656  & 0.0271 & 0.0769  \\
    PPID-WFANet  & 31.59 & 0.8829 & 9.9394 & 5.6134 & 30.64  & 0.8124 & 24.3899 & 11.1096 & 0.7120  & 0.0439 & 0.0938  \\ 
    LSAN-PanGAN  & 37.10 & 0.9684 & 7.7853 & 3.1186 & 33.77  & 0.9285 & 20.8042 & 7.7551 & 0.7600  & 0.0893 & 0.0587  \\
    LSAN-VBPN    & 35.19 & 0.9334 & 9.0635 & 3.7457 & 33.67  & 0.8476 & 19.4300 & 7.7413 & 0.7547  & 0.0351 & 0.0789  \\
    LSAN-WFANet  & 35.30 & 0.9497 & 7.6442 & 3.6943 & 32.65  & 0.8938 & 20.8059 & 8.6553 & 0.6559  & 0.1283 & 0.0912  \\
    PMNet-PanGAN & 33.66 & 0.9370 & 11.4521 & 4.5034 & 33.12 & 0.8947 & 21.0539 & 8.3400 & 0.7975 & 0.0630 & 0.0525 \\
    PMNet-VBPN   & 31.89 & 0.8878 & 10.3304 & 5.3827 & 32.03 & 0.8480 & 19.0317 & 9.3013 & 0.7552 & \textbf{0.0148} & 0.0849 \\
    PMNet-WFANet & 31.70 & 0.8936 & 10.8468 & 5.5127 & 31.58 & 0.8366 & 20.7981 & 9.7760 & 0.7134 & 0.0399 & 0.0947 \\
    EFN          & 40.55 & 0.9833 & 5.3147 & 2.2556 & 40.66  & 0.9792 & 15.8017 & 3.8981 & 0.8681  & 0.0614 & 0.0257  \\ \hline
    \textbf{Ours}            & \textbf{41.59} & \textbf{0.9849} & \textbf{5.2912}  & \textbf{1.9039}  & \textbf{41.97} & \textbf{0.9829} & \textbf{15.4414} & \textbf{3.3768}  & \textbf{0.8709} & 0.0623 & \textbf{0.0243}   \\ \hline
\end{tabular}%
}
\caption{Quantitative evaluation of the competing methods on the CAVE, Chikusei, and Real-world datasets. $\uparrow$ stands for the greater value the better, while $\downarrow$ stands for the smaller value the better. The best result of each quality index is marked in \textbf{bold}.}
\label{Quantitative evaluation}
\end{table*}

\section{Experiments}
\subsection{Datasets}
We employ two widely-used hyperspectral datasets, i.e., CAVE \cite{yasuma2010generalized}, Chikusei \cite{yokoya2016airborne}, for simulation experiments. Meanwhile, we employ the Real-world dataset \cite{EFN} for real-world experiment.

\noindent\textbf{CAVE Dataset} is a prominent benchmark for hyperspectral imaging provided by Columbia University' s Computer Vision Laboratory. This collection comprises 31 spectral channels ranging from 400 to 700~nm with a 10~nm sampling interval, each featuring a spatial resolution of $512 \times 512$ pixels. In our study, we extract a subset spanning the $12^{\text{th}}$ to the $27^{\text{th}}$ bands to produce a 16-band HSI. We apply independent channel-wise normalization by scaling pixel intensities relative to the maximum value within each band. For experimental evaluation, the 32 available images are split into a training set of 20 and a testing set of 12.

\noindent\textbf{Chikusei Dataset} is acquired over Chikusei City using the Hyperspec-VNIR-CIRIS sensor. It encompasses 128 valid spectral bands with a $2335 \times 2517$ spatial grid and a 2.5~m ground sampling distance. Adopting the band selection criteria used for CAVE, we construct a 16-band HSI for our experiments. The data intensity is normalized by a factor of $2^{14}$. To partition the dataset, we divide the original scene along the width: a $2335 \times 2117 \times 16$ sub-cube is designated for training, while a $2335 \times 400 \times 16$ strip is reserved for testing. The latter is further segmented into five $400 \times 400$ and one $335 \times 400$ test images by cropping along the height.

For both synthetic datasets, the mosaiced and PAN observations are generated from the reference HSIs through a standardized degradation model. Specifically, we apply an $8 \times 8$ SFA mosaicing pattern followed by $2 \times 2$ average downsampling to synthesize the mosaiced inputs. The corresponding PAN images are derived by convolving the HSIs with a predefined SRF, represented as $(1, 1, 2, 4, 8, 9, 10, 12, 16, 12, 10, 9, 7, 3, 2, 1) / 107$.

\noindent\textbf{Real-world dataset} contains 60 pairs of mosaiced and PAN images spanning 16 spectral bands (723--896~nm, $\sim$10~nm intervals). The mosaiced data possess a resolution of $1020 \times 1104$, whereas the PAN images feature a doubled resolution of $2040 \times 2208$. These samples are captured in outdoor environments under natural illumination, maintaining a sufficient distance to ensure far-field imaging conditions. Raw 12-bit pixel values are scaled to a $[0, 1]$ range via $2^{12}$ normalization. We allocate 50 image pairs for model training and 10 for performance validation. For qualitative visualization, pseudo-color images are generated by assigning the $13^{\text{th}}$, $8^{\text{th}}$, and $3^{\text{rd}}$ bands to the red, green, and blue channels, respectively.

\begin{figure*}[t]
\centering
  \includegraphics[width=0.98\textwidth]{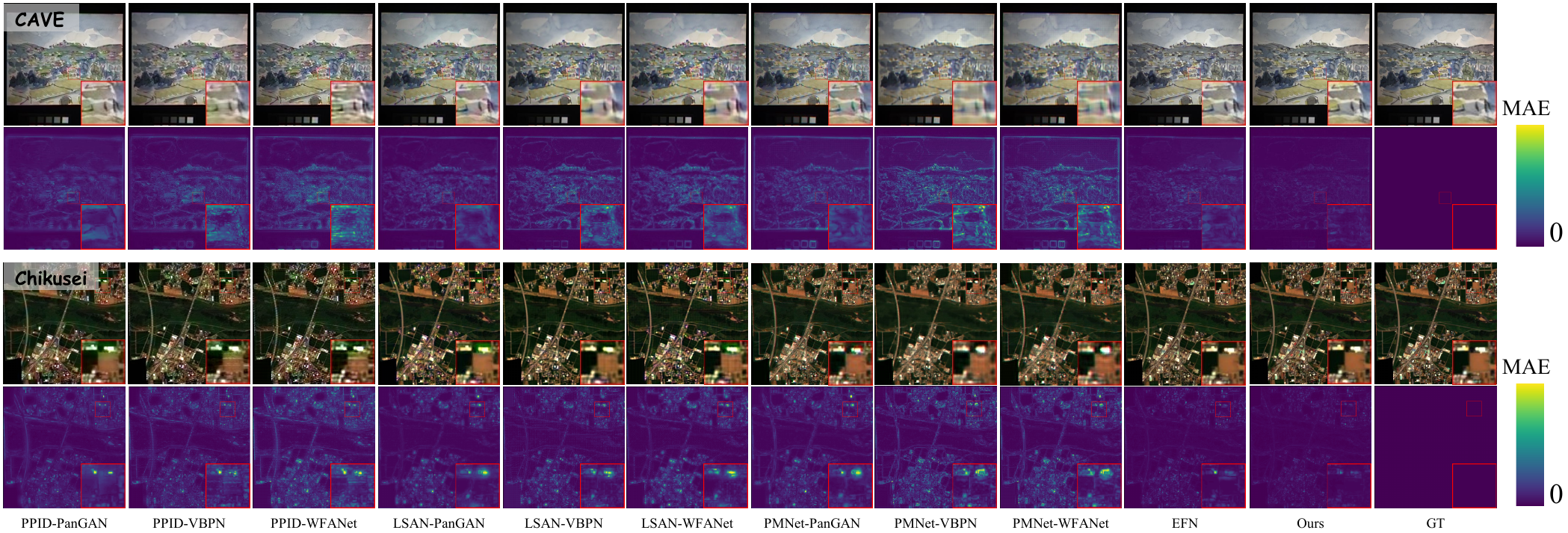}
  \caption{Qualitative evaluation of the competing methods on the CAVE and Chikusei datasets. Odd row: visualizations of the fusion results; even row: MAE maps.}
  \label{fig:simulated_results}
\end{figure*}

\begin{figure*}[h]
\centering
  \includegraphics[width=0.98\textwidth]{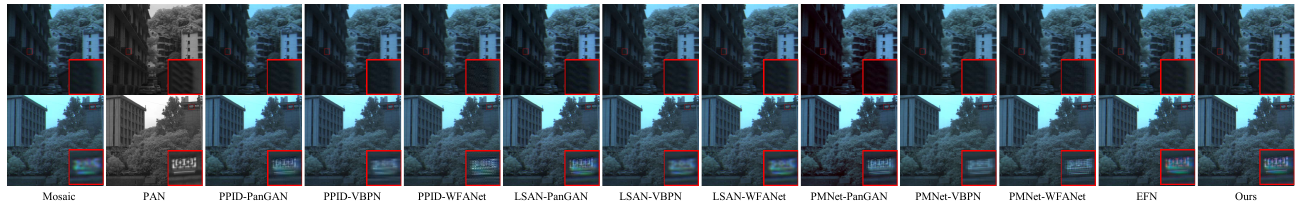}
  \caption{Qualitative evaluation of competing methods across two representative scenes from the Real-world dataset.}
  \label{fig:realworld_results}
\end{figure*}

\subsection{Experimental Settings}
\noindent\textbf{Implementation Details.}
All experiments were conducted on an NVIDIA RTX 4090 GPU. Since our training pipeline consists of two stages, we follow \cite{EFN} for all hyperparameter settings in the first pretraining stage. For the subsequent flow matching training phase across the three datasets, we crop the mosaiced and PAN images into $32 \times 32$ and $64 \times 64$ patches, respectively. The network parameters are optimized using the Adam optimizer with a fixed batch size of $16$. We apply specific learning rates to different modules: $10^{-4}$ for the flow matching model and $10^{-5}$ for the PAN degradation network. Furthermore, during the inference stage, the number of sampling steps for the flow matching process is set to $10$ to balance computational efficiency and generation quality.

\noindent\textbf{Competing methods.}
The task of mosaiced and PAN images fusion typically includes two sequential phases: spectral demosaicing followed by pansharpening. To establish a robust comparative framework, we evaluate three state-of-the-art demosaicing algorithms, namely PPID \cite{mihoubi2017multispectral}, LSAN \cite{LSAN}, and PMNet \cite{liu2025exploring}, which facilitate the recovery of full-spectrum data from raw mosaiced inputs. These are then coupled with three premier pansharpening techniques, namely PanGAN \cite{PanGAN}, VBPN \cite{VBPN}, and WFANet \cite{huang2025wavelet}, to enhance spatial detail using HR PAN image. This configuration yields nine distinct two-stage fusion pipelines. We also include EFN \cite{EFN} as a competing method for single-stage fusion strategy.


\noindent\textbf{Evaluation Metrics.} For simulated datasets with available ground truth, we employ four widely recognized metrics: PSNR, SSIM \cite{zhou2004image}, SAM \cite{SAM}, and ERGAS \cite{ERGAS}. For real-world scenarios lacking reference images, we utilize no-reference quality indices, specifically $D_\lambda$, $D_S$, and QNR \cite{QNR}. Notably, we adapt the $D_\lambda$ metric to better suit our problem by reshaping the raw mosaiced input into a LR hyperspectral cube format prior to computation. Unlike typical satellite-derived data, which benefit from extensive pre-processing, our real-world dataset is collected in raw format. Consequently, inherent sensor noise may decouple standard no-reference metrics from human visual perception. To address this discrepancy, we apply a $5 \times 5$ Gaussian blur with a standard deviation of $\sigma = 1$ to both mosaiced and PAN images before evaluation, ensuring a more stable and perceptually aligned assessment of fusion quality.

\subsection{Comparison experiments}
To demonstrate the superiority of our proposed framework, we perform extensive evaluations on simulated and Real-world datasets, benchmarking it against current state-of-the-art methods.

\noindent\textbf{Quantitative evaluation on the simulated datasets.} 
As summarized in Table~\ref{Quantitative evaluation}, our method consistently outperforms competing approaches on both simulated datasets. Specifically, on the CAVE dataset, it achieves a PSNR of 41.59 dB and an SAM of 5.2912 on the CAVE dataset, surpassing the second-best EFN (PSNR 40.55 dB, SAM 5.3147). A similar trend is observed on the Chikusei dataset, where our model leads with a PSNR of 41.97 dB and SAM of 15.4414. These results demonstrate that our method delivers superior hyperspectral reconstructions, effectively balancing spatial and spectral accuracy against strong baselines like EFN and LSAN-PanGAN.

\noindent\textbf{Qualitative evaluation on the simulated datasets.} Fig.~\ref{fig:simulated_results} displays the qualitative results of all competing methods on the CAVE and Chikusei datasets. As observed on the CAVE dataset, PPID-WFANet, PMNet-VBPN, and PMNet-WFANet exhibit noticeable artifacts, while LSAN-PanGAN, LSAN-VBPN, and LSAN-WFANet suffer from obvious spectral discrepancies. Although EFN yields a low MAE compared to the ground-truth HSI, it still exhibits significant spectral distortion. On the Chikusei dataset, the PPID-based variants, LSAN-PanGAN, and LSAN-WFANet show clear spectral distortion, whereas LSAN-VBPN and the PMNet-based variants introduce apparent spatial artifacts. Overall, our proposed method achieves the best performance in terms of both spectral and spatial fidelity, demonstrating its powerful reconstruction capabilities.

\begin{figure}[t]
\centering
  \includegraphics[width=0.98\linewidth]{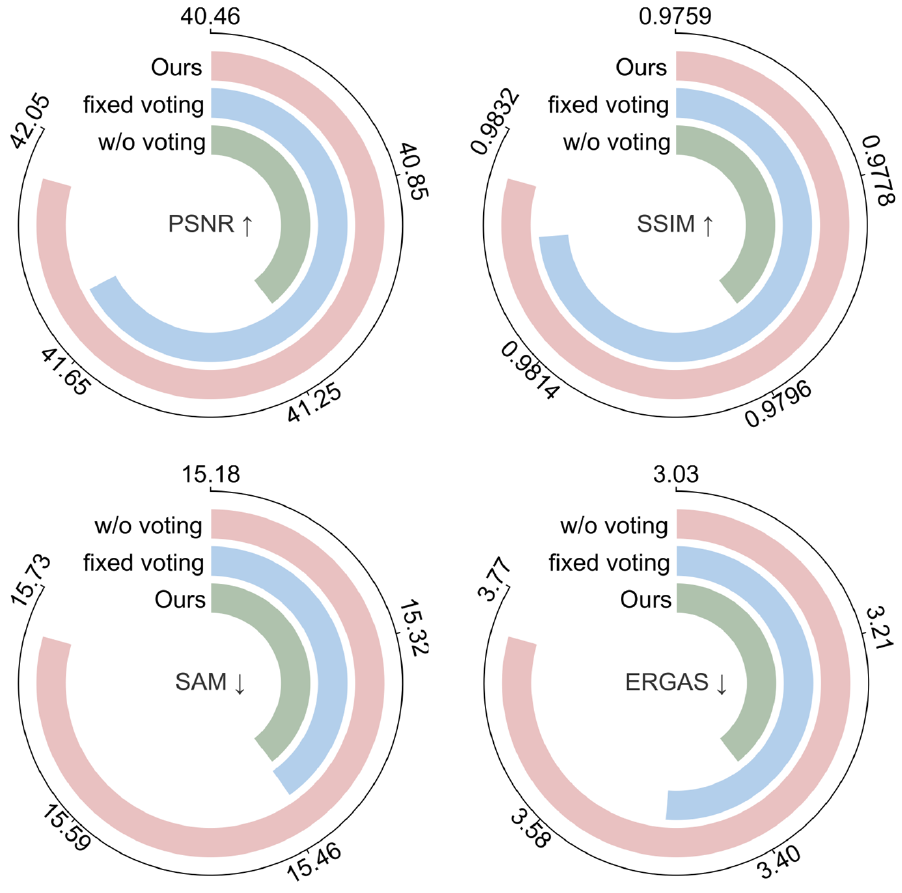}
  \caption{Ablation study on the random voting mechanism.}
  \label{fig:ablation_voting}
\end{figure}

\begin{figure}[t]
\centering
  \includegraphics[width=0.98\linewidth]{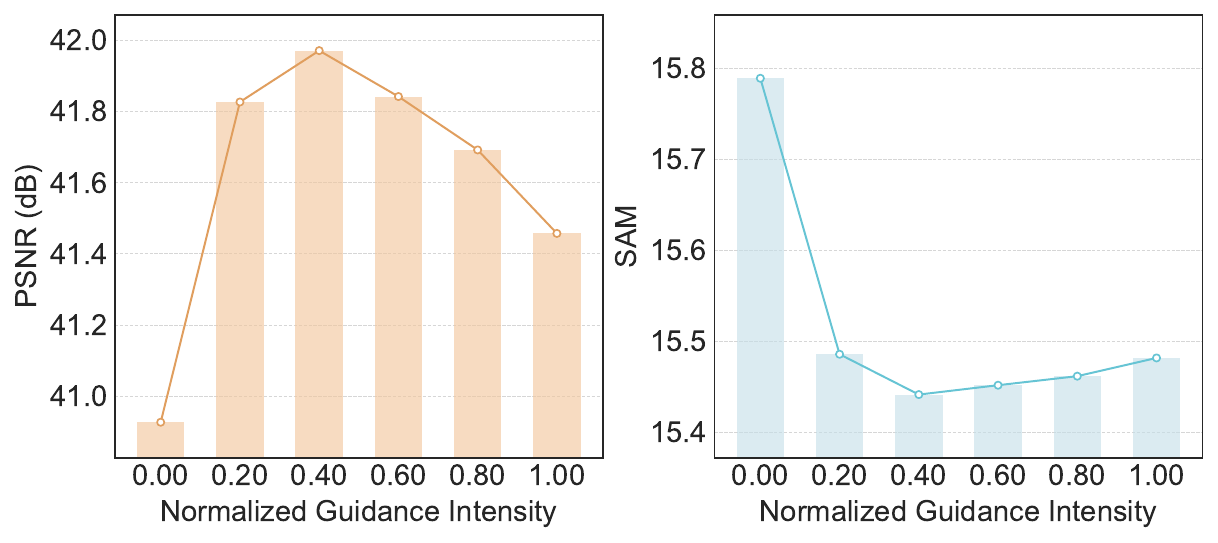}
  \caption{Ablation study on the effect of guidance intensity.}
  \label{fig:guidance_ablation}
\end{figure}

\noindent\textbf{Quantitative evaluation on the Real-world dataset.} As reported in Table~\ref{Quantitative evaluation}, the results on the Real-world dataset demonstrate the superior generalization capability of our method. Our approach achieves the highest QNR of 0.8709 and the lowest spatial distortion ($D_S$) of 0.0243, outperforming the second-best method, EFN. Although PMNet-VBPN yields a lower spectral distortion ($D_{\lambda}$ of 0.0148), our model maintains a highly competitive performance while providing significantly better overall reconstruction quality, as reflected by the leading QNR metric. These quantitative findings confirm that our proposed architecture is not only effective for simulated data but also highly reliable for real-world applications.

\noindent\textbf{Qualitative evaluation on the Real-world dataset.} Fig.~\ref{fig:realworld_results} depicts the qualitative results of all competing methods on the Real-world dataset. As observed, PPID-VBPN, PPID-WFANet, PMNet-PanGAN, PMNet-VBPN, and PMNet-WFANet exhibit severe artifacts, while LSAN-VBPN and LSAN-WFANet suffer from obvious blurring. Among these competing methods, PPID-PanGAN, LSAN-PanGAN, and EFN achieve a relative trade-off between the spatial and spectral fidelity. In contrast, our proposed method achieves the best performance in reconstructing the HR-HSI, demonstrating superiority over existing fusion methods. 

\subsection{Ablation study}
We conduct ablation experiments on the Chikusei dataset to evaluate the contributions of the main components of our framework.

\noindent\textbf{Impact of the random voting mechanism.} To evaluate the effectiveness of the proposed random voting mechanism, we compare our full model against two ablated variants: one without this mechanism and another adopting a fixed selection strategy (\emph{i.e.}, using the last $k$ checkpoints within a evaluation window). 
As shown in Fig.~\ref{fig:ablation_voting}, the proposed random voting strategy consistently outperforms both counterparts across all metrics. This improvement indicates that the random voting mechanism effectively refines the HR-HSI estimate, mitigating the propagation of errors from unreliable pseudo labels, and optimizing the training of the flowing matching toward improved spatial and spectral fidelity.
This superiority of random sampling over fixed selection stems from its enhanced robustness, which alleviates the local bias induced by successive checkpoints. 

\noindent\textbf{Impact of the conflict-free guidance.} In this part, we evaluate the impact of the gradient guidance intensity. For clarity, the original intensity parameter range $[0,50]$ is normalized to $[0,1]$. As shown in Fig.~\ref{fig:guidance_ablation}, introducing gradient guidance yields a substantial improvement over the zero-gradient baseline (intensity=$0$), demonstrating its effectiveness in enforcing spatial and spectral consistency.
The model reaches its peak performance at a normalized scale of $0.4$. At this optimal point, the gradient guidance provides sufficient spatial and spectral cues without overpowering the model's learned prior. However, beyond this point, performance gradually declines. This decline suggests that excessively large gradients push the sampling trajectory off the natural data manifold. Therefore, we adopt a moderate normalized intensity of $0.4$ to optimally balance explicit spectral and spatial consistency with the implicit generative prior.
 
\section{Conclusion}
In this study, we propose a generalizable and computationally efficient semi-supervised flow matching framework for mosaiced and PAN image fusion, overcoming the limitations of existing diffusion-based methods imposed by restrictive protocols or handcrafted assumptions. We introduce a novel and effective random voting mechanism that enables reliable training of the flow matching model on real-world data. During inference, we adopt a conflict-free gradient guidance strategy to achieve spatially and spectrally faithful HR-HSI reconstruction. Extensive experiments demonstrate that our method consistently outperforms state-of-the-art approaches in both quantitative metrics and visual quality, especially highlighting its strong potential for real-world applications.

\section{Acknowledgement}
This paper is supported by the National Natural Science Foundation of China (No. 62422108, No. 62576135, and No. 62221002), the Supported by Fujian Provincial Natural Science Foundation of China (No. 2024J01227), the State Key Laboratory of Ocean Engineering (Shanghai Jiao Tong University), and the State Key Laboratory of Spatial Datum (No. SKLSD2025-KF-18).






\bibliographystyle{ACM-Reference-Format}
\bibliography{main}








\end{document}